%% file: main.tex
\documentclass[runningheads]{llncs}
\usepackage[T1]{fontenc}
\usepackage{graphicx}
\usepackage{xcolor} 
\usepackage{caption}
\usepackage{subcaption}
\usepackage{svg}
\usepackage{graphicx}
 \usepackage{amsmath}

 \usepackage{tikz}
\usetikzlibrary{shapes.geometric, arrows.meta, positioning, calc, backgrounds, fit}

\definecolor{myblue}{RGB}{30, 144, 255}
\definecolor{mygreen}{RGB}{50, 205, 50}
\definecolor{myorange}{RGB}{255, 165, 0}
\definecolor{myred}{RGB}{220, 20, 60}
\definecolor{mypurple}{RGB}{147, 112, 219}

\usepackage{float}
\usepackage{multirow}
\usepackage[table]{xcolor}
\definecolor{gray2}{rgb}{0.9, 0.9, 0.9}
\usepackage{booktabs}
\usepackage[table]{xcolor}
\usepackage{makecell}
\usepackage{multirow}
\renewcommand{\arraystretch}{1.2}

\usepackage{amsmath}
\usepackage{amssymb}
\usepackage{caption}

\begin{document}
%
\title{Adaptive Scaffolding for Cognitive Engagement in an Intelligent Tutoring System}
%
\titlerunning{Adaptive Scaffolding for Cognitive Engagement}
%
\authorrunning{Tithi. et al.}
\author{Sutapa Dey Tithi\inst{1}
\and
Nazia Alam\inst{1}
\and 
Tahreem Yasir\inst{1}
\and
Yang Shi\inst{2}
\and
Xiaoyi Tian\inst{1}
\and
Min Chi\inst{1}
\and
Tiffany Barnes\inst{1}
}
\institute{North Carolina State University
\and
Utah State University
}

%
\maketitle              

\input{Latex/0_abstract}
\input{Latex/1_introduction}
\input{Latex/2_related_work}
\input{Latex/3_method}

\input{Latex/4_experiment}
\input{Latex/5_result}
\input{Latex/6_discussion}
\input{Latex/7_conclusion}



\bibliographystyle{splncs04}
\bibliography{myBibliography}

\end{document}

%% file: Latex/0_abstract.tex
\begin{abstract}
The ICAP framework defines four cognitive engagement levels: Passive, Active, Constructive, and Interactive, where increased cognitive engagement can yield improved learning. However, personalizing learning activities that elicit the optimal level of cognitive engagement remains a key challenge in intelligent tutoring systems (ITS). In this work, we develop and evaluate a system that adaptively scaffolds cognitive engagement by dynamically selecting worked examples in two different ICAP modes: \textit{active} Guided examples and \textit{constructive} Buggy examples. We compare Bayesian Knowledge Tracing ($BKT$) and Deep Reinforcement Learning ($DRL$) as adaptive methods against a non-adaptive baseline method for selecting example type in a logic ITS. Our experiment with 113 students demonstrates that both adaptive policies significantly improved student performance on test problems. $BKT$ yielded the largest improvement in posttest scores for low prior knowledge students, helping them catch up with their high prior knowledge peers, whereas $DRL$ yielded significantly higher posttest scores among high prior knowledge students. This paper contributes new insights into the complex interactions of cognitive engagement and adaptivity and their results on learning outcomes. 



\keywords{Intelligent Tutoring Systems \and Cognitive Engagement \and ICAP \and Deep Reinforcement Learning \and Bayesian Knowledge Tracing}
\end{abstract}

%% file: Latex/1_introduction.tex
\section{Introduction}
A key factor in effective scaffolding in intelligent tutoring systems (ITSs) is learners' cognitive engagement \cite{van2015effects}. The ICAP learning framework differentiates cognitive engagement into four hierarchical modes, in order of increasing engagement and learning: \textit{Passive}--receiving information, \textit{Active}--manipulating materials, e.g., copying steps or selecting from menus, \textit{Constructive}-- generating outputs beyond provided materials, e.g., self-explaining, and \textit{Interactive}-- collaborative knowledge construction \cite{chi2014icap,vanlehn2007tutorial,conati2000toward}. Previous studies demonstrate consistent learning improvements with increased engagement in diverse domains \cite{chi2014icap,mitrovic2019investigating,wiggins2017icap}, suggesting that ICAP provides a principled foundation for designing scaffolding. However, higher engagement levels may require sufficient prior knowledge or additional scaffolding within ITSs. 

ITSs commonly employ Problem Solving (PS), where students construct solutions independently, or Worked Examples (WE), where tutors present step-by-step solutions. While worked examples reduce cognitive load for novices, as expertise develops, detailed guidance designed for novices becomes redundant and can hinder performance, which is known as the \textit{expertise reversal effect} \cite{sweller2011cognitive,kalyuga2009expertise}. This effect has been documented across various domains, including mathematics \cite{kalyuga2001problem}, science \cite{rey2011expertise}, and procedural tasks \cite{oksa2010expertise}. Research suggests that more cognitively-engaging activities benefit high-expertise students, while explicit guidance helps low-expertise students \cite{chi2014icap}. This raises a key question: can learning be optimized by \textit{adaptively} scaffolding cognitive engagement based on evolving student knowledge or performance?

Prior work has demonstrated that data-driven approaches can effectively induce pedagogical policies for adaptive scaffolding. Bayesian Knowledge Tracing (BKT), which models student mastery based on observed performance, has been used widely to guide instructional decisions in tutoring systems \cite{corbett1994knowledge,yudelson2013individualized}. More recently, Deep Reinforcement Learning (DRL) has shown success in learning pedagogical policies that optimize student learning in adaptive environments \cite{wang2018reinforcement,sanz2020exploring,doroudi2019s}. While BKT provides interpretable, theory-grounded mastery estimates, DRL can capture complex temporal dependencies to optimize learning outcomes. Given these strengths and their successes in similar step-based tutoring contexts, we apply both approaches to adaptively scaffold cognitive engagement in an intelligent logic tutor.

In this work, we develop and evaluate a system that adaptively scaffolds cognitive engagement by dynamically selecting worked examples at two different ICAP engagement levels. Our tutor provides problem solving (PS) interleaved with two types of worked examples: \textit{Buggy} examples (requiring students to identify and fix errors, classified as \textit{constructive} engagement in ICAP) and \textit{Guided} examples (requiring students to complete missing justifications, classified as \textit{active} engagement in ICAP). We explore two adaptivity approaches: a BKT-based heuristic policy and a DRL policy, both evaluated against a random baseline in a classroom study with 113 students. We address the following research questions: 
\begin{itemize}
    \item RQ1: How do adaptive (BKT, DRL) and non-adaptive (Control) policies differ in their distribution of scaffolding types and time-on-task during training?
    \item RQ2: Do adaptive scaffolding policies improve posttest performance compared to the non-adaptive policy? 
    \item RQ3: Do the effects of scaffolding policy on posttest performance vary by prior knowledge level?
\end{itemize}


%% file: Latex/2_related_work.tex
\section{Background and Related Work}
Worked examples (WEs) have been widely studied as scaffolding tools in intelligent tutors \cite{renkl2002worked,mostafavi2015data}. While traditional WEs benefit novices by reducing cognitive load, they may hinder learners with higher expertise who must process redundant information \cite{kalyuga2009expertise}. Researchers have explored variations of WEs to address this limitation. In programming education, Parsons problems, which are partially worked examples, have been extensively explored and found to improve students' code writing abilities \cite{weinman2021improving,ericson2022parsons}. Parsons problems with subgoal labels demonstrated reduced difficulty and improved learning \cite{shabrina2023learning}. 

``Buggy'' or erroneous examples presenting incorrect solutions for students to identify and correct have also shown promise \cite{booth2013using,adams2014using}. However, the effectiveness of erroneous examples may depend heavily on prior knowledge since novices may lack the necessary schemas to identify errors, while advanced learners may benefit from the critical thinking required for debugging \cite{zhi2018exploring,mclaren2015delayed}. In this study, we designed two types of worked examples in our intelligent logic tutor using the ICAP framework \cite{chi2014icap}. The tutor provides Guided worked examples, which adapt the Parsons problem format to logic proofs and require \textit{active} engagement, and Buggy worked examples, which require \textit{constructive} engagement. This design allows us to investigate how adaptive selection between these engagement levels can accommodate learners with varying prior knowledge.

We use {Bayesian Knowledge Tracing (BKT)} and Deep reinforcement learning (DRL) as the adaptive methods. BKT is among the most widely studied student modeling approaches for tracing students' knowledge evolvement \cite{corbett1994knowledge}. BKT models student performance, continuously updating mastery estimates based on observed correct and incorrect responses. The model has demonstrated effectiveness across numerous student modeling tasks and remains foundational in tutoring systems \cite{yudelson2013individualized,reye2004student}. 

{Reinforcement Learning (RL)} provides a framework for inducing data-driven scaffolding policies that optimize cumulative learning outcomes \cite{de2018multi}. RL's flexibility in reward function design has made it increasingly popular in educational applications \cite{ju2021evaluating,sawyer2017balancing}. Deep reinforcement learning (DRL), which combines RL with neural networks, has shown particular promise for pedagogical policy induction in adaptive learning environments \cite{wang2018reinforcement,sanz2020exploring,abdelshiheed2023leveraging}. Deep Q-Networks approximate value functions without requiring explicit domain concepts, allowing application to complex state spaces characteristic of student modeling \cite{van2016deep}. Our study leverages these two popular adaptivity methods (BKT and DRL) to dynamically select problem types based on student knowledge. 


%% file: Latex/3_method.tex
\section{Methods}
\subsection{Tutor Overview}
\begin{figure*}[t]
\centering
\includegraphics[width=0.75\linewidth]{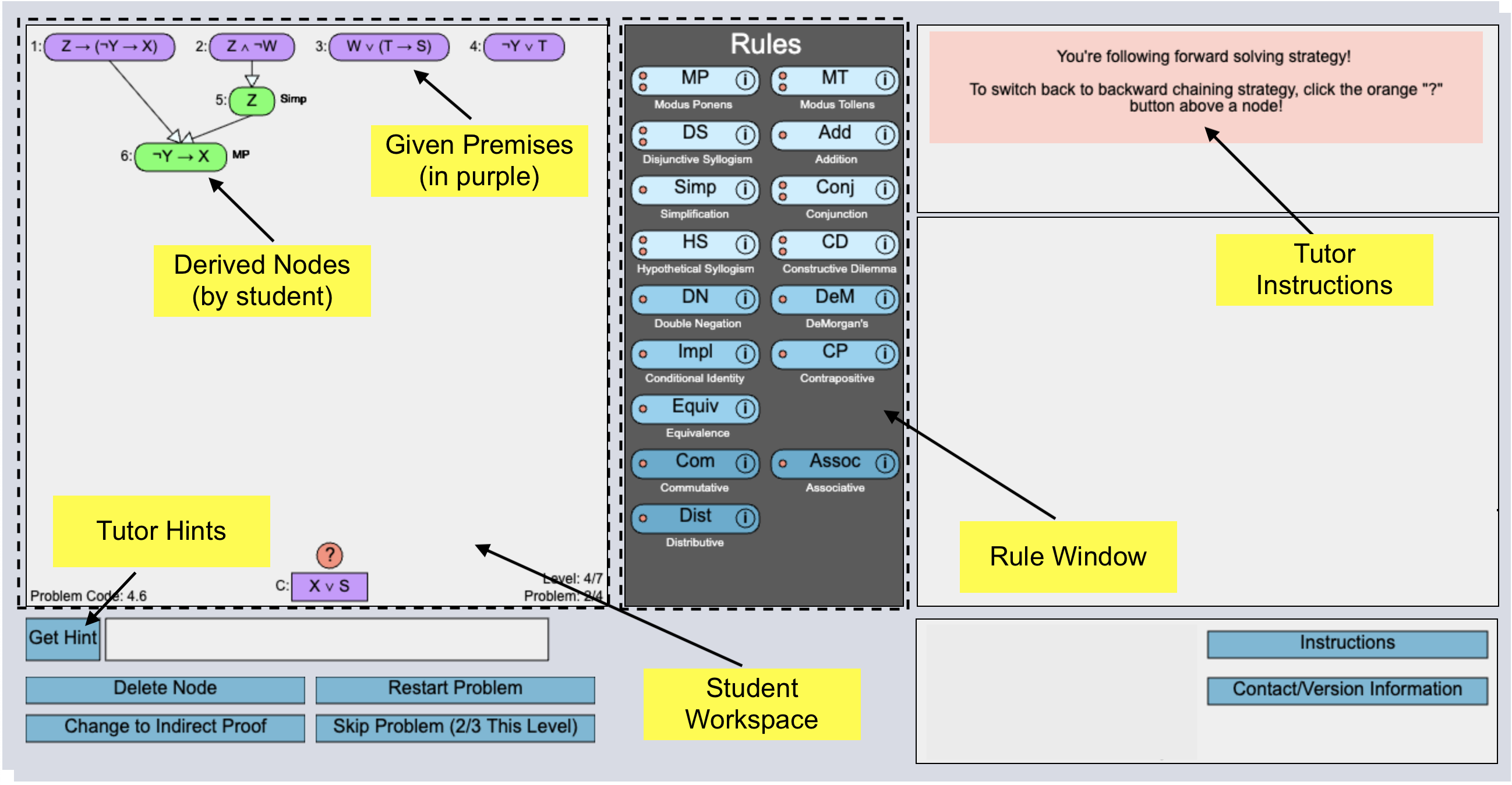}
\caption{The tutor interface with the integrated three-panel design: student workspace (left), domain rules (center), and instructions (right). Hints and feedback appear in the bottom-left corner.}
\label{fig:fig1}
\end{figure*}

Our ITS is used in an undergraduate discrete math course where students practice open-ended, multi-step propositional logic problem solving. Students interact with a three-panel interface (Figure~\ref{fig:fig1}). The left panel provides the primary workspace where students construct proofs by deriving logical statements represented as nodes. The center panel contains the rule buttons with their descriptions. The right panel displays contextual instructions. Students justify derived statements by selecting a rule and connecting 1--2 parent nodes via edges. The tutor includes four sections (introduction, pretest, training, and posttest) spanning seven levels. The introduction comprises the first two problems shown as (passive) worked examples (WEs) in level 1, helping students become acquainted with the interface. The next two problems in level 1, presented as problem-solving (PS), are used as the pretest to assess prior student knowledge. The training session consists of five ordered levels (levels 2--6) of increasing difficulty, each with four problems. The first three problems in each training level are either WEs or PS. The fourth problem on each level serves as a level-end test problem. The posttest level (level 7) consists of six test problems. It is important to note that the posttest was designed to be more challenging than the pretest, incorporating two isomorphic problems (7.1-7.2) to compare with the pretest, and transfer problems (7.3-7.6) to assess deeper understanding and generalization. A student’s problem score is a composite score (0--100) combining three normalized metrics: {rule accuracy}, {solution length}, and {problem completion time}, which ranks a solution based on how accurate, optimal, and fast it is \cite{shabrina2023learning,abdelshiheed2023leveraging}. Normalized learning gain is calculated using the average problem scores on the pre- and posttest problems with the equation $\text{NLG} = \frac{\text{posttest score} - \text{pretest score}}{\sqrt{100 - \text{pretest score}}}$ \cite{shabrina2023learning}.

\subsection{ICAP-Inspired Problem Types}
We operationalize the ICAP engagement modes through three problem representations (Figure~\ref{fig:fig1}, Figure~\ref{fig:problem_types}):
\begin{figure*}[t]
    \centering    
    \begin{subfigure}[b]{0.45\textwidth}
        \centering
        \includegraphics[width=\textwidth]{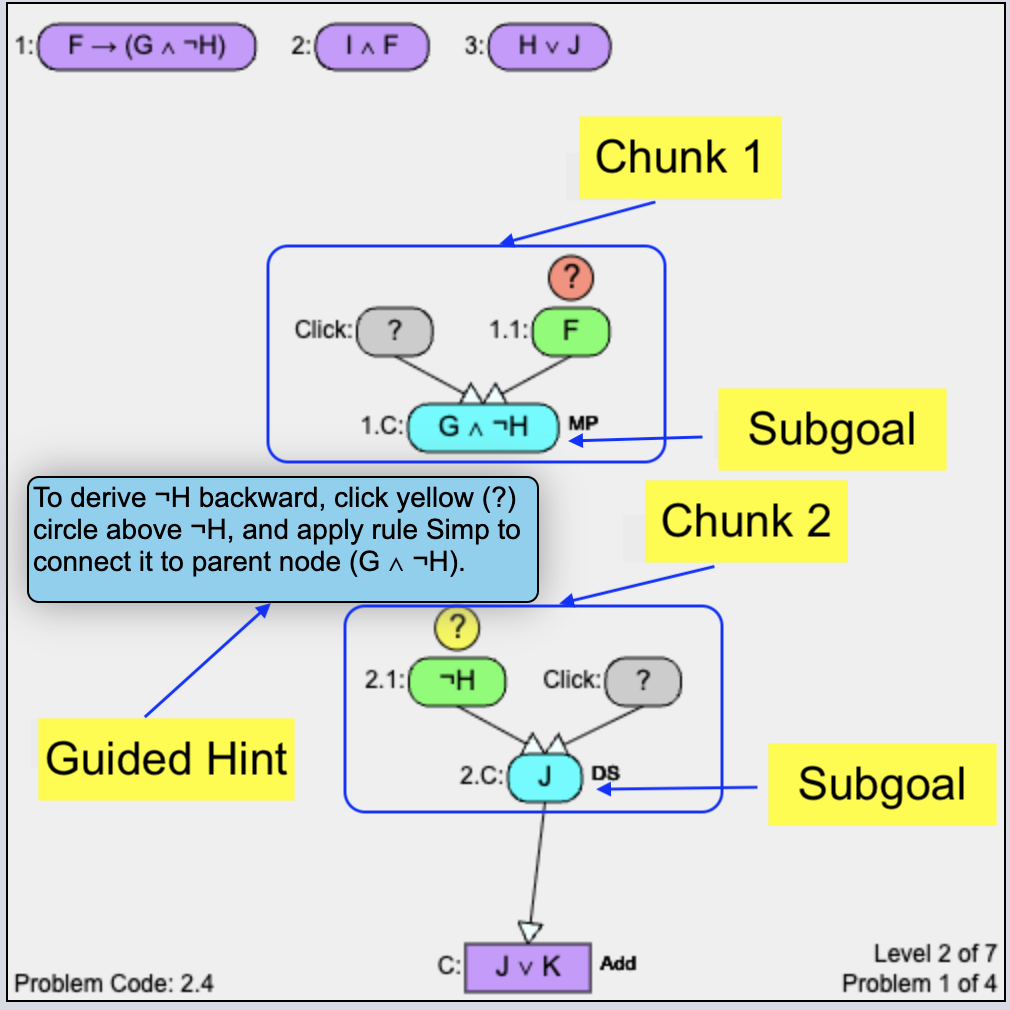}
        \caption{Guided Example (Guided): Partial solution with missing connections}
        \label{fig:gpp}
    \end{subfigure}
    \hspace{0.03\textwidth}
    \begin{subfigure}[b]{0.45\textwidth}
        \centering
        \includegraphics[width=\textwidth]{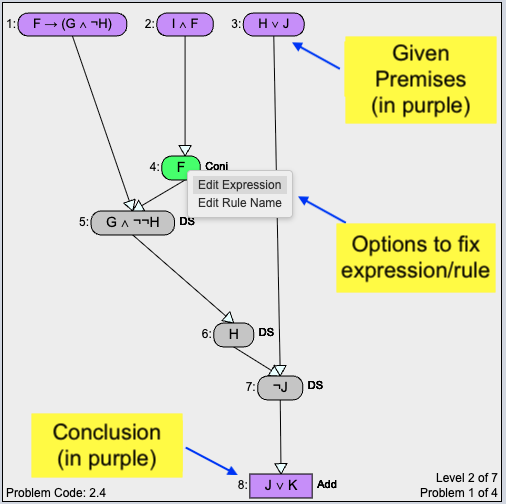}
        \caption{Buggy Example (Buggy): Partial solution with bugs inserted}
        \label{fig:buggy}
    \end{subfigure}

    \caption{The tutor interfaces for two example types designed based on the ICAP framework}
    \label{fig:problem_types}
\end{figure*}

\textbf{Problem Solving (PS)} requires students to independently derive all proof steps, demanding the highest, interactive cognitive engagement (Figure~\ref{fig:fig1}). The PS mode can be considered to be Interactive in the ICAP framework, since students receive explicit accuracy feedback on each step; students may also ask for hints during PS training problems. 

\textbf{Guided Worked Examples} present partially completed solutions organized into meaningful subgoals (Figure~\ref{fig:gpp}), following Renkl's ``meaningful building blocks'' principle \cite{renkl2002worked}. Students complete missing justifications where step-specific hints address the ``rationale gap'' in traditional worked examples \cite{renkl2002worked}. This design maintains low cognitive load through subgoal structure representing the \textit{active} engagement level in ICAP. In a Guided example, a partial solution with all the statement nodes needed to complete a proof is presented, and step-specific hints are attached to guide the students. Students must apply appropriate rules to complete missing edges between nodes. As shown in Figure~\ref{fig:gpp}, statement 2.1: $-H$ can be derived from 1.C: $G \land -H$ using rule Simplification. To complete this step, students click on the yellow question mark above 2.1, choose the rule Simplification, and click on statement 1.C to create an edge from 1.C to 2.1. The student repeats this process until all the  nodes are connected.

\textbf{Buggy Worked Examples} present solutions containing expert-designed errors (``bugs'') that students must identify and correct (Figure~\ref{fig:buggy}). Students click on nodes to either fix the logic statement or its associated rule, then submit corrections for verification. This format engages students in understanding the given solution, finding bugs, and typing their own solution elements, representing the \textit{constructive} engagement level in ICAP. Figure \ref{fig:buggy} shows an intermediate solution state of a Buggy example, where the given nodes 1, 2, and 3 are shown at the top, and the conclusion node 8 ($J \lor K$) is shown at the bottom; the given nodes and conclusion are always correct. When the student corrects an expression, it turns green (Node 4 ($F$)). When the student clicks node 4 again, the student can select the option `Edit Rule Name' to replace the given rule (`Conj') with the correct rule (`Simp'). The student repeats this process until all bugs are corrected (and the number of remaining bugs is always visible).

\subsection{Adaptive Policy Induction}
\textbf{BKT-based Heuristic Policy.}
This policy routes students to either Buggy or Guided examples based on mastery estimates derived from Bayesian Knowledge Tracing (BKT). Students demonstrating lower mastery receive Guided examples with step-by-step scaffolding; students demonstrating higher mastery receive Buggy examples requiring higher cognitive engagement. We adopt a data-driven method to apply BKT for problem solving when the domain has rules but no knowledge components~\cite{mostafavi2017evolution}. The $ruleScore_i$ for a given rule $i$ is initialized with the parameters: the initial probability of knowing a rule $p(L_0) = 0.01$, the probability of transitioning from unlearned to learned $p(T) = 0.01$, the probability of answering a question correctly when a rule is not learned $p(G) = 0.3$, and the probability of answering a question incorrectly when a rule is learned $p(S) = 0.1$. After each rule application, mastery estimates are updated following standard BKT inference equations~\cite{mostafavi2017evolution}. After each problem, rule-specific scores ($ruleScore_i$) are compared against threshold values ($ruleThreshold_i$) computed from prior tutor deployments ($n = 721$ students across seven course sections). The scores from the past semesters for each rule score were averaged after each problem completion and set as $ruleThreshold_i$. 
For each student, every $rule_i$ is assigned a positive $scoreSign_i$ value if the rule score is above $ruleThreshold_i$. Each $rule_i$ receives a negative $scoreSign_i$ value if the score is below the $ruleThreshold_i$, meaning that they have not yet shown the same level of proficiency in rule application as past students. This positive or negative value is weighted based on the rule priority (1.0 for rules required in the current problem; 0.5 otherwise). The weighted $scoreSign_i$ values are summed, and the sign of the sum determines whether a student is directed to the Buggy example or the Guided example for the next problem.

\textbf{DRL Policy.}
We employed an offline Double Deep Q-Network (DDQN) to induce the DRL policy \cite{van2016deep}. The adaptive scaffolding task is formalized as a Markov Decision Process where \textit{State (S)} is the 74-dimensional normalized feature vector capturing: (1) \textit{mastery features}, including rule-specific accuracy and mastery estimates; (2) \textit{temporal features}, including problem completion time, time since last hint request, and session duration; and (3) \textit{help-seeking features}, including hint requests per problem, hint utilization rates, and error frequencies. \textit{Action (A)} is a Buggy Example, Guided Example, or Problem Solving. \textit{Reward (R)} captures both learning outcomes and time efficiency earning: 
\begin{equation}
\text{Reward} = \textit{TestScore} \cdot (1 - \textit{ProblemTime})
\end{equation}
where TestScore is the average score on level-end and posttest problems, and ProblemTime is the normalized completion time for the current problem. DDQN decouples action selection and evaluation to mitigate overestimation bias \cite{van2016deep}. This provides improved stability and convergence during training. The network architecture consists of fully connected layers with ReLU activations, followed by a linear output layer for the three actions. The policy was trained offline on prior data from 103 students comprising 1,570 state-action-reward transitions across training levels. After hyperparameter tuning, we selected the model with the lowest mean squared error loss. The deployed policy had fully connected layers ($\{64, 128, 64\}$ units), learning rate $10^{-3}$, discount factor $\gamma=0.99$, batch size 100, and target network synchronization every 50 steps.


%% file: Latex/4_experiment.tex
\subsection{Experiment Setup and Participants}
We conducted a controlled between-subjects study with three training conditions: $Control$, $BKT$, and $DRL$. (1) For $Control$: Problem type was selected uniformly at random from problem-solving (PS), Guided examples, or Buggy examples. (2) For $BKT$: First PS or worked example was selected uniformly at random. If a worked example was selected, the specific type (Guided or Buggy) was then determined by the BKT-based heuristic policy according to the student's estimated knowledge state; (3) For $DRL$: PS, Guided, or Buggy examples were adaptively assigned using the DRL policy. \footnote{Note that all decisions were overridden as needed by a requirement that all levels have at least one PS training problem type to ensure independent practice. In other words if the policy chose Guided, Buggy, Guided, the third choice would have been replaced by PS.} All conditions followed the same tutor structure (pretest, training, posttest) and had access to identical problem content; only the selection mechanism for problem types differed.

The tutor was deployed as a lab assignment in an undergraduate discrete math course at a U.S. public university ($N = 113$). We did not collect course-specific demographics; however, as Discrete Math is mandatory for CS majors, we report demographics of the 2021--22 CS graduating class as an approximation: 83\% men, 17\% women; 58\% white, 18.5\% Asian, 3\% Hispanic/Latin, 2\% Black/African American, 9\% other races, 9.5\% international students. This study received IRB exemption, and only authorized researchers could access participant data. Students were assigned to conditions after completing pretest problems, using stratified random sampling to ensure an even distribution of pretest scores across conditions. We only analyze students who completed all 7 levels: 36 in $Control$, 35 in $BKT$, and 42 in $DRL$.

\subsection{Analysis}
To investigate our research questions, we analyzed the selected problem types, pretest, level-end test, and posttest scores, and their composite parts (accuracy, optimality, and time). We performed Shapiro-Wilk tests for each metric in Table~\ref{tab:overall_results}, which indicated our data were non-normal ($p < 0.05$). We therefore adopted a combination of regression and non-parametric statistical tests (Kruskal-Wallis and Mann-Whitney with Bonferroni correction for continuous outcomes; chi-square for categorical distributions) \cite{weisstein2004bonferroni}. All reported p-values for pairwise comparisons are Bonferroni-adjusted. We reported probability-based effect sizes ($A$) with 95\% confidence intervals ($CI$) for this analysis \cite{ruscio2008probability}. 

%% file: Latex/5_result.tex
\section{Results}
We report our results by research question, first describing differences in problem types and time across conditions (RQ1), learning outcomes (RQ2), and then impacts on low versus high prior knowledge groups (RQ3).

\subsection{RQ1: Scaffolding Distribution and Time-on-task}

\begin{table}[t]
\centering
\caption{Time by each tutor section (hours, Mean (SD)) and problem type distribution during training across three conditions}
\label{tab:time_problem_distribution}
\small
\setlength{\tabcolsep}{5pt}
\renewcommand{\arraystretch}{0.9}
\begin{tabular}{@{}lccc|ccc@{}}
\toprule
 & \multicolumn{3}{c}{Time (hours)} & \multicolumn{3}{c}{Training Problem Type} \\
\cmidrule(lr){2-4} \cmidrule(lr){5-7}
Group (N) 
& Training 
& Level-End 
& Posttest 
& PS 
& Buggy 
& Guided \\
\midrule
\multirow{2}{*}{$Control$ (35)}
& \multirow{2}{*}{1.84 (0.8)}
& \multirow{2}{*}{1.53 (0.9)}
& \multirow{2}{*}{1.18 (0.5)}
& 241 & 203 & 164 \\
& & & & (39.6\%) & (33.4\%) & (27.0\%) \\
\hline

\multirow{2}{*}{$BKT$ (36)}
& \multirow{2}{*}{1.64 (0.9)}
& \multirow{2}{*}{1.57 (0.8)}
& \multirow{2}{*}{1.09 (0.9)}
& 277 & 201 & 161 \\
& & & & (43.3\%) & (31.5\%) & (25.2\%) \\
\hline
\multirow{2}{*}{$DRL$ (42)}
& \multirow{2}{*}{1.78 (0.9)}
& \multirow{2}{*}{1.47 (0.9)}
& \multirow{2}{*}{1.17 (0.8)}
& 244 & 29 & 416 \\
& & & & (35.4\%) & (4.2\%) & (60.4\%) \\
\bottomrule
\end{tabular}
\end{table}

First, we compare the time students spent in each section of the tutor and the distribution of problem types across conditions (Table~\ref{tab:time_problem_distribution}, Figure~\ref{fig1:problemType}). Overall time was comparable across conditions, with students in all three groups spending approximately five hours with the tutor. Average time spent per problem on each problem type was not significantly different across conditions ($p > .05$). A chi-square test showed no significant difference in the distribution of problem types between $Control$ and $BKT$ groups ($\chi^2(2) = 1.77$, $p = 1.00$), with about 29-44\% PS, 31-33\% Buggy, and 25-27\% Guided. However, students in $DRL$ group encountered significantly different distribution from both $Control$ ($\chi^2(2) = 131.2$, $p < .001$), and $BKT$ ($\chi^2(2) = 143.1$, $p < .001$), with 35\% PS, only 4\% Buggy, and 60\% Guided. 

\subsection{RQ2: Learning Outcomes}
Next, we compare students' performance in posttest problems. There were no significant differences in pretest scores ($Control$ (mean): 70.2, $BKT$ (mean): 68.0, $DRL$ (mean): 70.7; $p > 0.05$). We performed a mixed-effect regression analysis, with training conditions as the independent variable, posttest score as the dependent variable, and problem ID as a random intercept (to control for differences across problems). The results demonstrated a significant association between the training conditions and posttest score ($p < 0.001$). As shown in Table~\ref{tab:overall_results}, both adaptive conditions significantly outperformed the $Control$ (mean $= 65.7$) on posttest score. The $BKT$ group (mean $= 72.3$) showed significant improvement over $Control$ ($A = .58$, 95\% CI [.52, .63], $p = .005$)\footnote{Reported $p-values$ are Bonferroni corrected for multiple pairwise comparisons.}. Similarly, the $DRL$ group (mean $= 72.5$) also demonstrated significant improvement ($A = .58$, 95\% CI [.53, .64], $p = .002$). The two adaptive conditions did not differ significantly from each other. 
 
Beyond overall scores, students in $BKT$ completed posttest problems significantly faster than students in $Control$ ($Control$ (mean) = 11.8 minutes per problem, $BKT$ (mean) = 10.8 minutes, $p$ = 0.02, $A = .58$, 95\% CI [.53, .63]). $DRL$ group had marginally lower problem completion time than the $Control$ group in posttest problems ($Control$ (mean) = 11.8 minutes, $DRL$ (mean) = 10.7 minutes, $p$ = 0.06, $A = .56$, 95\% CI [.51, .61]). The $DRL$ group had marginally more optimal solutions with fewer steps than the $Control$ group in posttest problems ($Control$ (mean) = 8.5 steps, $DRL$ (mean) = 7.7 steps, $p$ = 0.06\textcolor{black}{, $A = .56$, 95\% CI [.51, .61])}\footnote{Note that in this domain, shorter solutions with fewer steps are more optimal, and shorter times are more time efficient.}. There were no significant differences in rule accuracy across groups ($Control$ (mean): 72.1, $BKT$ (mean): 74.8, $DRL$ (mean): 74.7; $p > 0.05$).

\begin{table}[t]
\centering
\caption{Average per-problem posttest performance across three conditions (Mean (SD)); Scores are higher for high rule accuracy, low time, and low steps. [Note: \textsuperscript{*} represents $p$-value<0.05, and \textsuperscript{$\dagger$} represents marginal significance.]}
\label{tab:overall_results}
\resizebox{\columnwidth}{!}{%
\begin{tabular}{@{}lccccc@{}}
\toprule
\textbf{Group (N)} & \textbf{Score} & \textbf{NLG}  & \textbf{Rule Acc.} & \textbf{Time (min)} & \textbf{Steps} \\
\midrule
Control (36) & 65.7 (22.7) & 0.34 (0.3) & 72.1 (21.3) & 11.8 (15.8) & 8.5 (3.8) \\
BKT (35) & 72.3 (21.8)\textsuperscript{*} & 0.50 (0.2)& 74.8 (21.5) & 10.8 (21.1)\textsuperscript{*} & 7.9 (3.3) \\
DRL (42) & 72.5 (21.5)\textsuperscript{*} & 0.46 (0.2)\textsuperscript{$\dagger$} & 74.7 (22.1) & 10.7 (20.6)\textsuperscript{$\dagger$} & 7.7 (3.0)\textsuperscript{$\dagger$} \\
\bottomrule
\end{tabular}
}
\end{table}
 
\subsection{RQ3: Effects of Adaptive Policies by Prior Knowledge}
\begin{figure*}[t]
    \centering    
    \begin{subfigure}[b]{0.48\textwidth}
        \centering
        \includegraphics[width=\textwidth]{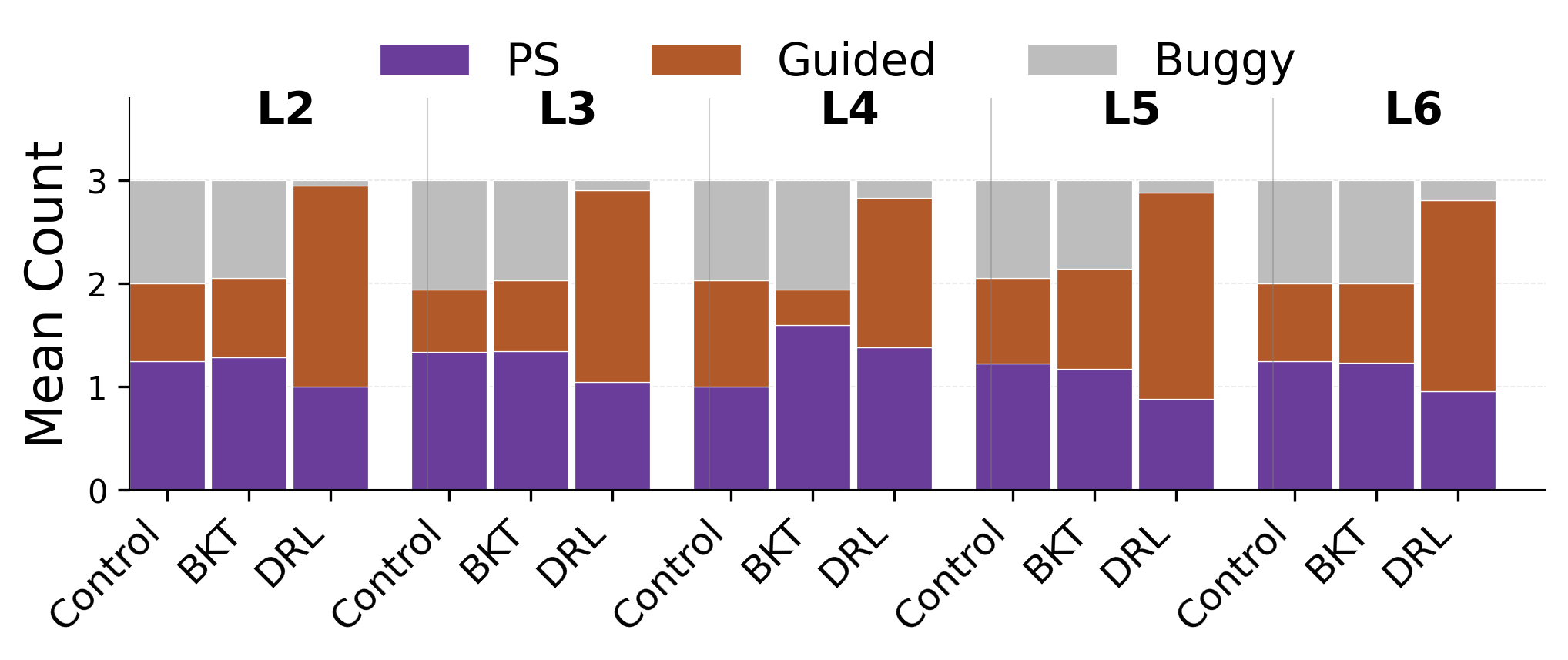}
        \caption{Distribution of problem types (PS, Guided, Buggy) in the first three training problems per level across conditions}
        \label{fig1:problemType}
    \end{subfigure}
    \hspace{0.01\textwidth}
    \begin{subfigure}[b]{0.48\textwidth}
        \centering
        \includegraphics[width=\textwidth]{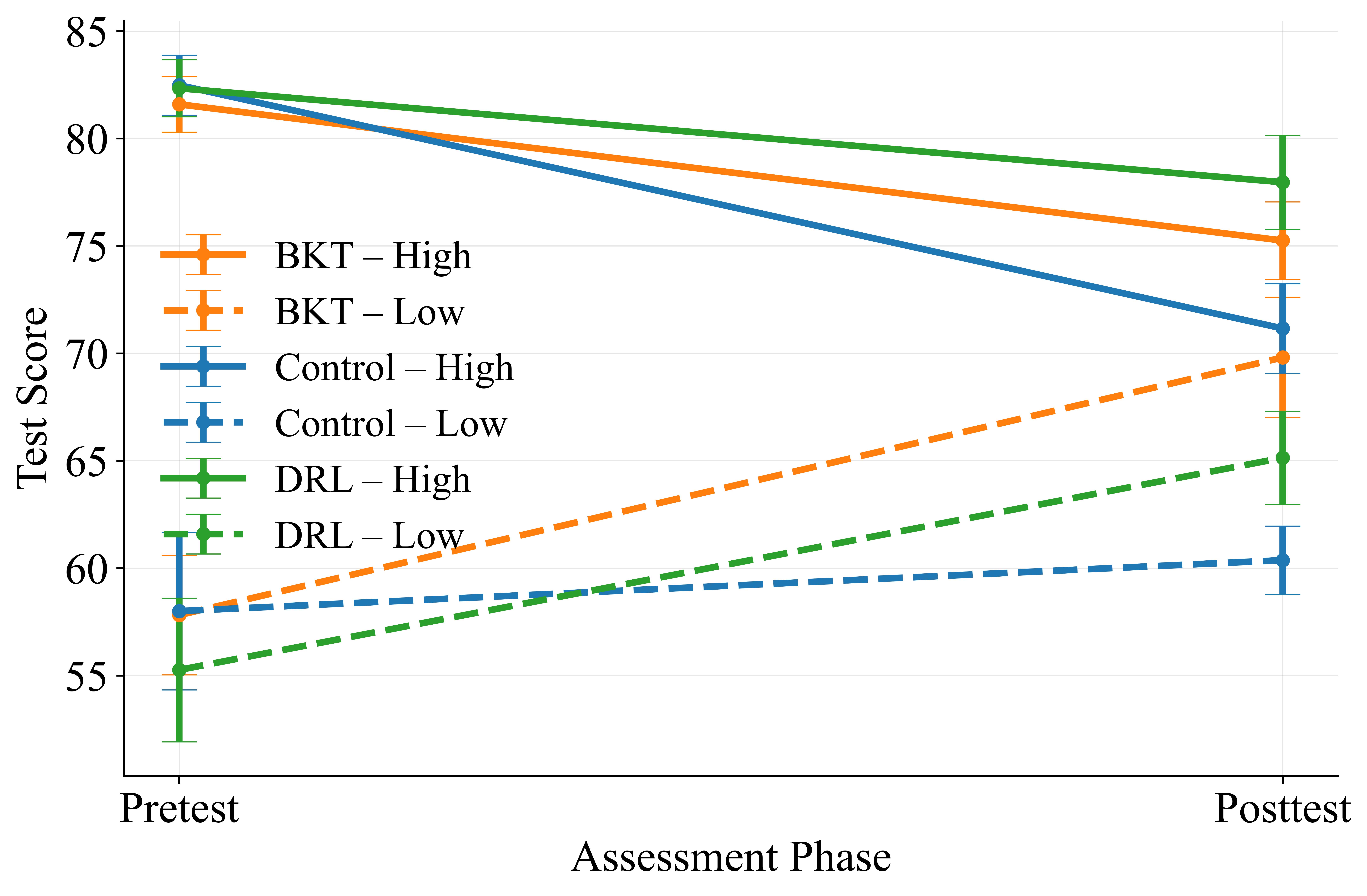}
        \caption{Mean pretest to posttest scores by prior knowledge}
        \label{fig:low_high}
    \end{subfigure}
    \caption{Problem type distribution and learning outcomes across conditions.}
    \label{fig:problem_score}
\end{figure*}

Next, we investigated whether our training interventions showed an aptitude-treatment interaction, with low prior performance groups benefiting more than high prior performance groups \cite{snow1991aptitude}. We defined High and Low prior-knowledge groups based on a median split on pretest scores to partition the students into six Prior-Condition subgroups: High-Control ($n=18$), Low-Control ($n=18$), High-BKT ($n=20$), Low-BKT ($n=15$), High-DRL ($n=20$), and Low-DRL ($n=22$) (Figure~\ref{fig:low_high}). No significant differences were found on pretest performance between corresponding subgroups
We fit a mixed-effects regression model on \emph{posttest problem scores}. We modeled Condition and Prior Knowledge as fixed effects, and problem ID as a random intercept. Low-Control was the reference group (mean $= 60.4$). Students in Low-BKT significantly outperformed Low-Control ($\beta = 9.4$, $SE=2.7$, $z=3.4$, 
$p = .001$), while students in Low-DRL demonstrated marginal increase ($\beta = 4.8$, $SE=2.8$, $z=1.7$, 
$p = .08$). Students in High-Control scored significantly higher than students in Low-Control (
$p < .001$). 

The interaction terms were not statistically significant (High-BKT: 
$\beta = -5.3$, $SE=3.9$, $z=-1.4$, 
$p = .18$; High-DRL: 
$\beta = 2.0$, $SE=3.8$, $z=0.6$, 
$p = 0.59$), indicating that the advantage of adaptive conditions over $Control$ did not differ significantly by prior knowledge level. To further examine condition differences within the High prior knowledge subgroup, we conducted follow-up comparisons: students in High-DRL achieved a significantly higher posttest score than students in High-Control ($\beta = 6.81$, $SE = 2.6$, $z = 2.6$, $p = 0.008$), while students in High-BKT did not ($\beta = 4.09$, $SE = 2.9$, $z = 1.4$, $p = .16$). 



These relationships between prior knowledge and condition are shown in Figure~\ref{fig:low_high}. Examining this figure illustrates that the adaptive conditions $BKT$ and $DRL$ had similar (parallel) effects on the Low and High prior knowledge groups. However, the $Control$ condition failed to improve the Low prior knowledge group (with a mostly flat slope from pre to post) while leaving the High prior knowledge group less prepared for the (much harder) posttest than the adaptive conditions. All training policies narrowed the achievement gap from pretest to posttest based on prior knowledge, with reductions of 55.9\% $Control$, 52.6\% $DRL$, and 77.1\% for $BKT$. The achievement gap at a test is the difference in mean scores between the High and Low prior knowledge groups $Test_{gap} = Test_{High} - Test_{Low}$.
$Control$ reduced the gap by 55.9\% from 24.5 to 10.8 pre to post ($Pre_{gap}= 24.5 = Pre_{High} 82.5 - Pre_{Low} 58.0$; $Post_{gap} = 10.8 = Post_{High} 71.2 - Post_{Low} 60.4$; $p<0.001$). $DRL$ achieved a similar 52.6\% reduction, from 27 to 12.8, ($Pre_{gap} = 27 = Pre_{High} 82.3 - Pre_{Low} 53.3$; $Post_{gap} = Post_{High} 77.9 - Post_{Low} 65.1$; $p<0.001$). $BKT$ achieved the largest gap reduction of 77.1\%, narrowing the difference from 23.8 to 5.4 from pre to post ($Pre_{gap} = 23.8 = Pre_{High} 81.6 - Pre_{Low} 57.8$; $Post_{gap} = 5.4 = Post_{High} 75.3 - Post_{Low} 69.8$; $p<0.001$). 

%% file: Latex/6_discussion.tex
\section{Discussion}
Overall, the two adaptive policies induced different scaffolding strategies. The $DRL$ reward function was designed to optimize both learning and time efficiency, which might have led $DRL$ to favor \textit{active} engagement (Guided examples) over \textit{interactive} but time consuming PS and \textit{constructive} engagement (Buggy examples), as shown in Figure \ref{fig:problem_types}. This finding can be interpreted through the lens of the ICAP framework and cognitive load theory \cite{chi2014icap,sweller2011cognitive}. Guided examples may offer a more efficient path to mastery than either PS or Buggy examples by reducing extraneous cognitive load while maintaining germane load \cite{sweller2011cognitive}. PS takes more time, and students may not solve problems optimally, while Buggy problems are constructive while ensuring that students eventually see the most optimal solutions with the fewest steps. Buggy examples impose additional cognitive demands over Guided ones but may promote deeper processing \cite{adams2014using}. Prior research on buggy examples has shown that while they can enhance learning, their effectiveness depends on learner readiness; novices may struggle without sufficient prior knowledge to recognize and repair errors \cite{booth2013using,mclaren2015delayed}. $BKT$ produced problem type distributions similar to $Control$, but $BKT$ tended to first assign Guided examples for students with lower rule knowledge estimates, and then followed up with Buggy examples if and when students improved. This sequence may have helped struggling students to build and reinforce domain knowledge early in the tutor. 
As can be seen in Figure \ref{fig:problem_types}, $Control$ assigned more Guided and fewer PS problems in Level L4 than $BKT$. This may have occurred because of the random nature of the assignment of PS problems and could have contributed to the differences between the Low-BKT and Low-Control groups. Level L5 is the most difficult level in the tutor, introducing 4 new rules for practice. It may be that introducing more PS problems helped prepare the Low-BKT group for the future L5 learning.

Both adaptive conditions significantly outperformed the $Control$ group on posttest performance, with comparable effect sizes ($A = .58$). This finding aligns with prior research demonstrating the benefits of adaptive instruction in ITS \cite{vanlehn2011relative}, while further showing that adaptation based on \textit{cognitive engagement levels} yields significantly better posttest performance than non-adaptive instruction. The comparable overall performance between $BKT$ and $DRL$ suggests that the benefit may stem primarily from adaptation itself rather than the specific algorithmic approach in this study. 

Our findings have several implications for the design of adaptive learning systems. This study demonstrates that varying the \textit{type} of cognitive engagement through different representations and interactions can be an effective adaptation strategy. The ICAP framework provides a theoretically grounded basis for designing such adaptations. $BKT$'s mastery estimates allowed us to interpret why students received particular scaffolding types. By its nature, $DRL$'s learned policy is less interpretable. In educational settings where teachers and learners benefit from understanding system decisions, the interpretability gap between deep learning and knowledge modeling can be a meaningful consideration. At the same time, defining the ICAP level of a learning activity is non-trivial. Both Guided and Buggy examples provide feedback, and if the student reflected upon the feedback to make a subsequent change to their approach, either type could become more interactive \cite{chi2014icap}. 
$DRL$'s strong preference for \textit{active} Guided examples over \textit{interactive} PS and \textit{constructive} Guided examples, while yielding similar learning outcomes, raises questions about when constructive or interactive engagement is worth its cognitive cost. Constructive activities such as self-explanation, buggy error correction, and knowledge integration are theorized to produce more transferable learning \cite{chi2014icap,chi1994eliciting}, yet they require greater cognitive effort and may not always translate to immediate performance gains. $DRL$'s avoidance of interactive and constructive levels of engagement may also represent a missed pedagogical opportunity. Interactive and constructive engagement may lack immediate time efficiency gains but promote deeper learning that our deep learning rewards may not have fully captured. 

%% file: Latex/7_conclusion.tex
\section{Conclusion}
This work demonstrates how adaptive policies can be used to personalize the \textit{type} of cognitive engagement students experience during problem-solving. By operationalizing ICAP engagement levels through worked example types, we provide a theoretically motivated implementation of cognitive engagement-oriented adaptation in an intelligent logic tutor, though it represents one of many possible approaches. 

The study has several limitations. The DRL training data came from a single semester, thus may not generalize across learner populations. Moreover, we implemented these interventions with one tutor. Our design and findings should be replicated in other problem-solving domains, such as science, math, or programming, to determine their generalizability. Although Guided and Buggy examples were designed to elicit \textit{active} and \textit{constructive} engagement, respectively, further research is needed to fully understand the cognitive effort required by Guided and Buggy examples, as well as students' perceived difficulty with each example type, since perceived difficulty may differ from objective cognitive demands and could influence both engagement and learning outcomes. Future work should explore adaptive methods that reward engagement or productive struggle rather than performance alone, and also examine the extent to which constructive engagement yields benefits beyond immediate posttest performance.


%% file: myBibliography.bib
@article{mostafavi2017evolution,
  title={Evolution of an intelligent deductive logic tutor using data-driven elements},
  author={Mostafavi, Behrooz and Barnes, Tiffany},
  journal={International Journal of Artificial Intelligence in Education},
  volume={27},
  number={1},
  pages={5--36},
  year={2017},
  publisher={Springer}
}

@article{snow1991aptitude,
  title={Aptitude-treatment interaction as a framework for research on individual differences in psychotherapy.},
  author={Snow, Richard E},
  journal={Journal of consulting and clinical psychology},
  volume={59},
  number={2},
  pages={205},
  year={1991},
  publisher={American Psychological Association}
}

@incollection{kalyuga2009expertise,
  title={The expertise reversal effect},
  author={Kalyuga, Slava},
  booktitle={Managing cognitive load in adaptive multimedia learning},
  pages={58--80},
  year={2009},
  publisher={IGI Global Scientific Publishing}
}

@article{kalyuga2001problem,
  title={When problem solving is superior to studying worked examples.},
  author={Kalyuga, Slava and Chandler, Paul and Tuovinen, Juhani and Sweller, John},
  journal={Journal of educational psychology},
  volume={93},
  number={3},
  pages={579},
  year={2001},
  publisher={American Psychological Association}
}

@article{oksa2010expertise,
  title={Expertise reversal effect in using explanatory notes for readers of Shakespearean text},
  author={Oksa, Annishka and Kalyuga, Slava and Chandler, Paul},
  journal={Instructional Science},
  volume={38},
  number={3},
  pages={217--236},
  year={2010},
  publisher={Springer}
}

@incollection{sweller2011cognitive,
  title={Cognitive load theory},
  author={Sweller, John},
  booktitle={Psychology of learning and motivation},
  volume={55},
  pages={37--76},
  year={2011},
  publisher={Elsevier}
}

@article{chi1994eliciting,
  title={Eliciting self-explanations improves understanding},
  author={Chi, Michelene TH and De Leeuw, Nicholas and Chiu, Mei-Hung and LaVancher, Christian},
  journal={Cognitive science},
  volume={18},
  number={3},
  pages={439--477},
  year={1994},
  publisher={Elsevier}
}

@article{chi2014icap,
  title={The ICAP framework: Linking cognitive engagement to active learning outcomes},
  author={Chi, Michelene TH and Wylie, Ruth},
  journal={Educational psychologist},
  volume={49},
  number={4},
  pages={219--243},
  year={2014},
  publisher={Taylor \& Francis}
}

@inproceedings{weinman2021improving,
  title={Improving Instruction of Programming Patterns with Faded Parsons Problems},
  author={Weinman, Nathaniel and Fox, Armando and Hearst, Marti A},
  booktitle={Proceedings of the 2021 CHI Conference on Human Factors in Computing Systems},
  pages={1--4},
  year={2021}
}

@article{shabrina2023learning,
  title={Learning Problem Decomposition-Recomposition with Data-Driven Chunky Parsons Problems within an Intelligent Logic Tutor.},
  author={Shabrina, Preya and Mostafavi, Behrooz and Tithi, Sutapa Dey and Chi, Min and Barnes, Tiffany},
  journal={International Educational Data Mining Society},
  year={2023},
  publisher={ERIC}
}

@article{booth2013using,
  title={Using example problems to improve student learning in algebra: Differentiating between correct and incorrect examples},
  author={Booth, Julie L and Lange, Karin E and Koedinger, Kenneth R and Newton, Kristie J},
  journal={Learning and Instruction},
  volume={25},
  pages={24--34},
  year={2013},
  publisher={Elsevier}
}

@article{adams2014using,
  title={Using erroneous examples to improve mathematics learning with a web-based tutoring system},
  author={Adams, Deanne M and McLaren, Bruce M and Durkin, Kelley and Mayer, Richard E and Rittle-Johnson, Bethany and Isotani, Seiji and Van Velsen, Martin},
  journal={Computers in Human Behavior},
  volume={36},
  pages={401--411},
  year={2014},
  publisher={Elsevier}
}

@article{mclaren2015delayed,
  title={Delayed learning effects with erroneous examples: a study of learning decimals with a web-based tutor},
  author={McLaren, Bruce M and Adams, Deanne M and Mayer, Richard E},
  journal={International Journal of Artificial Intelligence in Education},
  volume={25},
  number={4},
  pages={520--542},
  year={2015},
  publisher={Springer}
}

@inproceedings{mostafavi2015data,
  title={Data-driven worked examples improve retention and completion in a logic tutor},
  author={Mostafavi, Behrooz and Zhou, Guojing and Lynch, Collin and Chi, Min and Barnes, Tiffany},
  booktitle={International conference on artificial intelligence in education},
  pages={726--729},
  year={2015},
  organization={Springer}
}

@article{renkl2002worked,
  title={Worked-out examples: Instructional explanations support learning by self-explanations},
  author={Renkl, Alexander},
  journal={Learning and instruction},
  volume={12},
  number={5},
  pages={529--556},
  year={2002},
  publisher={Elsevier}
}

@article{ericson2022parsons,
  title={Parsons problems and beyond: Systematic literature review and empirical study designs},
  author={Ericson, Barbara J and Denny, Paul and Prather, James and Duran, Rodrigo and Hellas, Arto and Leinonen, Juho and Miller, Craig S and Morrison, Briana B and Pearce, Janice L and Rodger, Susan H},
  journal={Proceedings of the 2022 working group reports on innovation and technology in Computer Science education},
  pages={191--234},
  year={2022}
}

@article{vanlehn2007tutorial,
  title={When are tutorial dialogues more effective than reading?},
  author={VanLehn, Kurt and Graesser, Arthur C and Jackson, G Tanner and Jordan, Pamela and Olney, Andrew and Ros{\'e}, Carolyn P},
  journal={Cognitive science},
  volume={31},
  number={1},
  pages={3--62},
  year={2007},
  publisher={Wiley Online Library}
}

@article{conati2000toward,
  title={Toward computer-based support of meta-cognitive skills: A computational framework to coach self-explanation},
  author={Conati, Cristina and Vanlehn, Kurt},
  journal={International Journal of Artificial Intelligence in Education},
  volume={11},
  pages={389--415},
  year={2000}
}

@article{weisstein2004bonferroni,
  title={Bonferroni correction},
  author={Weisstein, Eric W},
  journal={https://mathworld. wolfram. com/},
  year={2004},
  publisher={Wolfram Research, Inc.}
}

@article{ruscio2008probability,
  title={A probability-based measure of effect size: robustness to base rates and other factors.},
  author={Ruscio, John},
  journal={Psychological methods},
  volume={13},
  number={1},
  pages={19},
  year={2008},
  publisher={American Psychological Association}
}

@article{corbett1994knowledge,
  title={Knowledge tracing: Modeling the acquisition of procedural knowledge},
  author={Corbett, Albert T and Anderson, John R},
  journal={User modeling and user-adapted interaction},
  volume={4},
  number={4},
  pages={253--278},
  year={1994},
  publisher={Springer}
}

@inproceedings{van2016deep,
  title={Deep reinforcement learning with double q-learning},
  author={Van Hasselt, Hado and Guez, Arthur and Silver, David},
  booktitle={Proceedings of the AAAI conference on artificial intelligence},
  volume={30},
  number={1},
  year={2016}
}

@article{wang2018reinforcement,
  title={Reinforcement learning in a pomdp based intelligent tutoring system for optimizing teaching strategies},
  author={Wang, Fangju},
  journal={International Journal of Information and Education Technology},
  volume={8},
  number={8},
  pages={553--558},
  year={2018}
}

@inproceedings{de2018multi,
  title={Multi-step reinforcement learning: A unifying algorithm},
  author={De Asis, Kristopher and Hernandez-Garcia, J and Holland, G and Sutton, Richard},
  booktitle={Proceedings of the AAAI conference on artificial intelligence},
  volume={32},
  number={1},
  year={2018}
}

@inproceedings{ju2021evaluating,
  title={Evaluating critical reinforcement learning framework in the field},
  author={Ju, Song and Zhou, Guojing and Abdelshiheed, Mark and Barnes, Tiffany and Chi, Min},
  booktitle={International conference on artificial intelligence in education},
  pages={215--227},
  year={2021},
  organization={Springer}
}

@inproceedings{sanz2020exploring,
  title={Exploring the impact of simple explanations and agency on batch deep reinforcement learning induced pedagogical policies},
  author={Sanz Ausin, Markel and Maniktala, Mehak and Barnes, Tiffany and Chi, Min},
  booktitle={International conference on artificial intelligence in education},
  pages={472--485},
  year={2020},
  organization={Springer}
}

@article{van2015effects,
  title={The effects of scaffolding in the classroom: support contingency and student independent working time in relation to student achievement, task effort and appreciation of support},
  author={Van de Pol, Janneke and Volman, Monique and Oort, Frans and Beishuizen, Jos},
  journal={Instructional Science},
  volume={43},
  number={5},
  pages={615--641},
  year={2015},
  publisher={Springer}
}

@article{rey2011expertise,
  title={The expertise reversal effect: cognitive load and motivational explanations.},
  author={Rey, G{\"u}nter Daniel and Buchwald, Florian},
  journal={Journal of Experimental Psychology: Applied},
  volume={17},
  number={1},
  pages={33},
  year={2011},
  publisher={American Psychological Association}
}

@inproceedings{mitrovic2019investigating,
  title={Investigating the effect of adding nudges to increase engagement in active video watching},
  author={Mitrovic, Antonija and Gordon, Matthew and Piotrkowicz, Alicja and Dimitrova, Vania},
  booktitle={International conference on artificial intelligence in education},
  pages={320--332},
  year={2019},
  organization={Springer}
}

@article{wiggins2017icap,
  title={The ICAP active learning framework predicts the learning gains observed in intensely active classroom experiences},
  author={Wiggins, Benjamin L and Eddy, Sarah L and Grunspan, Daniel Z and Crowe, Alison J},
  journal={AERA Open},
  volume={3},
  number={2},
  pages={2332858417708567},
  year={2017},
  publisher={SAGE Publications Sage CA: Los Angeles, CA}
}

@article{doroudi2019s,
  title={Where’s the reward? a review of reinforcement learning for instructional sequencing},
  author={Doroudi, Shayan and Aleven, Vincent and Brunskill, Emma},
  journal={International Journal of Artificial Intelligence in Education},
  volume={29},
  number={4},
  pages={568--620},
  year={2019},
  publisher={Springer}
}

@inproceedings{sawyer2017balancing,
  title={Balancing learning and engagement in game-based learning environments with multi-objective reinforcement learning},
  author={Sawyer, Robert and Rowe, Jonathan and Lester, James},
  booktitle={International Conference on Artificial Intelligence in Education},
  pages={323--334},
  year={2017},
  organization={Springer}
}

@inproceedings{yudelson2013individualized,
  title={Individualized bayesian knowledge tracing models},
  author={Yudelson, Michael V and Koedinger, Kenneth R and Gordon, Geoffrey J},
  booktitle={International conference on artificial intelligence in education},
  pages={171--180},
  year={2013},
  organization={Springer}
}

@article{reye2004student,
  title={Student modelling based on belief networks},
  author={Reye, Jim},
  journal={International Journal of Artificial Intelligence in Education},
  volume={14},
  number={1},
  pages={63--96},
  year={2004},
  publisher={SAGE Publications Sage UK: London, England}
}

@inproceedings{abdelshiheed2023leveraging,
  title={Leveraging deep reinforcement learning for metacognitive interventions across intelligent tutoring systems},
  author={Abdelshiheed, Mark and Hostetter, John Wesley and Barnes, Tiffany and Chi, Min},
  booktitle={International Conference on Artificial Intelligence in Education},
  pages={291--303},
  year={2023},
  organization={Springer}
}

@article{vanlehn2011relative,
  title={The relative effectiveness of human tutoring, intelligent tutoring systems, and other tutoring systems},
  author={VanLehn, Kurt},
  journal={Educational psychologist},
  volume={46},
  number={4},
  pages={197--221},
  year={2011},
  publisher={Taylor \& Francis}
}

@inproceedings{zhi2018exploring,
  title={Exploring instructional support design in an educational game for K-12 computing education},
  author={Zhi, Rui and Lytle, Nicholas and Price, Thomas W},
  booktitle={Proceedings of the 49th ACM technical symposium on computer science education},
  pages={747--752},
  year={2018}
}
